\documentclass[conference]{IEEEtran}
\IEEEoverridecommandlockouts
\usepackage{algorithm}
\usepackage{algpseudocode}
\usepackage{graphicx}
\usepackage{array}
\usepackage[table,xcdraw]{xcolor} 
\usepackage{cite}
\usepackage{amsmath,amssymb,amsfonts}
\usepackage{textcomp}
\usepackage{makecell}
\usepackage{adjustbox}
\usepackage{subcaption}   % For subfigures and captions
\usepackage{float}        % For controlling figure placement
\usepackage{comment}
\usepackage[letterpaper, top=0.75in, left=1in, right=1in, bottom=1.1in]{geometry}

\def\BibTeX{{\rm B\kern-.05em{\sc i\kern-.025em b}\kern-.08em
    T\kern-.1667em\lower.7ex\hbox{E}\kern-.125emX}}

\begin{document}

\title{EcoFlight: Finding Low-Energy Paths Through Obstacles for Autonomous Sensing Drones\\
}

\author{Jordan Leyva, Nahim J. Moran Vera, Yihan Xu, Adrien Durasno, Christopher U. Romero,\\ Tendai Chimuka, Gabriel O. Huezo Ramirez, Ziqian Dong, and Roberto Rojas-Cessa
\thanks{J. Leyva, N. J. Mora Vera, A. Durasno, T. Chimuka, C.U. Romero, G.O. Ramirez,and R. Rojas-Cessa are with the Department of Electrical and Computer Engineering, New Jersey Institute of Technology, Newark, NJ, 07102.\protect \\
E-mail: rojas@njit.edu. \newline
Y. Xu and Z. Dong are with the Department of Electrical and Computer Engineering, New York Institute of Technology, New York,
NY, 10023.\protect\\
Email: ziqian.dong@nyit.edu}
}

\begin{comment}

\author{\IEEEauthorblockN{
%1\textsuperscript{st} 
Jordan Leyva}
\IEEEauthorblockA{
\textit{Electrical and Computer Engineering} \\
\textit{New Jersey Institute of Technology}\\
Newark, NJ, USA
%jl2326@njit.edu
}
\and
\IEEEauthorblockN{
%2\textsuperscript{nd} 
Nahim J. Moran Vera}
\IEEEauthorblockA{\textit{Computer Science} \\
\textit{New Jersey Institute of Technology}\\
Newark, NJ, USA
%njm58@njit.edu
}
\and
\IEEEauthorblockN{
%3\textsuperscript{rd} 
Adrien Durasno}
\IEEEauthorblockA{\textit{Electrical and Computer Engineering} \\
\textit{New Jersey Institute of Technology}\\
Newark, NJ, USA \\
%ad267@njit.edu
}
\and
\IEEEauthorblockN{
%4\textsuperscript{th} 
Christopher U. Romero}
\IEEEauthorblockA{\textit{Humanities and Social Sciences} \\
\textit{New Jersey Institute of Technology}\\
Newark, NJ, USA \\
%cur@njit.edu
}
\and
\IEEEauthorblockN{
%5\textsuperscript{th} 
Gabriel O. Huezo Ramirez}
\IEEEauthorblockA{\textit{Electrical and Computer Engineering} \\
\textit{New Jersey Institute of Technology}\\
Newark, NJ, USA \\
%gr48@njit.edu
}
\and
\IEEEauthorblockN{
%6\textsuperscript{th} 
Ziqian Dong}
\IEEEauthorblockA{\textit{Electrical and Computer Engineering} \\
\textit{New York Institute of Technology}\\
New York, NY, USA \\
%ziqian.dong@nyit.edu
}
\and
\IEEEauthorblockN{
%7\textsuperscript{th} 
Roberto Rojas-Cessa}
\IEEEauthorblockA{\textit{Electrical and Computer Engineering} \\
\textit{New Jersey Institute of Technology}\\
Newark, NJ, USA \\
%rcessa@njit.edu
}
}

\end{comment}

\maketitle

\begin{abstract}
%The role of drones in environmental sensing continues to grow, as they enable access to remote and difficult-to-reach areas, reduce costs, and lower the risk of injury to surveying personnel. 

%Most path planning schemes for drone flight automation assume that flying from one way point to another is simple and straight forward. However, there may be obstacles on the way and the flight path may need to be carefully found where energy efficiency must be considered in addition to collision avoidance. Therefore, we propose an energy-efficient path-finding algorithm, called the EcoFlight, that finds the path with the lowest energy expenditure in 3D space with obstacles in this paper. To realize this scheme,  we model the energy consumption of the drone based on the drone propulsion system and flying through a 3D space. We perform extensive evaluations and compare EcoFlight with schemes that may aim for a direct flight and one that targets the shortest distance. After extenuating simulations in various 3D environments that include various obstacle densities, our results show that EcoFlight finds paths with lower energy consumption than comparable planning algorithms, especially in areas with high obstacle density. 

%Most path planning schemes for drone flight automation assume that traveling between waypoints is simple and straightforward. In reality, flying through obstacles may require careful path selection, as energy efficiency must be considered in addition to collision avoidance. 

Obstacle avoidance path planning for uncrewed aerial vehicles (UAVs), or drones, is rarely addressed in most flight path planning schemes, despite obstacles being a realistic condition. Obstacle avoidance can also be energy-intensive, making it a critical factor in efficient point-to-point drone flights. To address these gaps, we propose EcoFlight, an energy-efficient pathfinding algorithm that determines the lowest-energy route in 3D space with obstacles. The algorithm models energy consumption based on the drone’s propulsion system and flight dynamics. We conduct extensive evaluations, comparing EcoFlight with direct-flight and shortest-distance schemes. The simulation results across various obstacle densities show that EcoFlight consistently finds paths with lower energy consumption than comparable algorithms, particularly in high-density environments. We also demonstrate that a suitable flying speed can further enhance energy savings.

%We not only identify paths with the lowest energy consumption but also highlight the role of the flying speed in minimizing energy consumption.
\end{abstract}


%--------------------------
%--------------------------
%--------------------------

\section{Introduction}
\label{sec:intro}

Drones, or Uncrewed Aerial Vehicles (UAVs), are increasingly used for environmental monitoring, search and rescue, and infrastructure inspection. %As wildfires, droughts, water pollution, and floods intensify due to climate change, drones have been proven valuable in recent events and are frequently deployed to monitor water quality in lakes, rivers, and reservoirs \cite{McDonald2019, hamzah2024drone}. 
We have witnessed their increasing deployment in recent disaster response and water quality monitoring in lakes, rivers, and reservoirs \cite{McDonald2019, hamzah2024drone}. However, their operation time is limited by the battery life. To maximize effectiveness, drones must optimize energy use to extend operational duration while ensuring a safe return to base. While propulsion efficiency contributes to energy savings, planning paths that minimize energy use remains a major challenge—particularly in obstacle-rich environments such as forests and cities.

Previous research has explored several methods for drone navigation, including A*, Dijkstra’s algorithm, genetic algorithms, and machine learning-based techniques \cite{hong2020least, li2018universal, shen2022multidepot}. These approaches aim to improve flight efficiency while minimizing computational requirements. However, many face challenges with real-time adjustments, multi-drone coordination, and adapting to dynamic environmental conditions such as wind, weather changes, and obstacles \cite{hong2020least, shivgan20233d}.

Although multiple strategies have been proposed for reducing energy consumption, less attention has been given to effectively integrating energy models into path-planning algorithms. To address this gap, we propose EcoFlight, an energy-efficient path planning algorithm based on A* that identifies routes with minimal energy expenditure. Unlike traditional A*, EcoFlight incorporates factors such as altitude changes, flight speed, and acceleration to dynamically select optimal paths. Rather than relying on complex AI training, EcoFlight retains the simplicity and reliability of A* through deterministic estimations, enabling improved efficiency in complex environments.

We evaluated EcoFlight through simulations and compared its performance with alternative path-finding algorithms, including direct path, distance-based A*, Rise and Traverse obstacle avoidance, and obstacle-agnostic direct flight. The results show that EcoFlight consistently outperforms distance-based approaches and is highly efficient in a high-density obstacle environment. %, particularly as obstacle density increases.

The remainder of this paper is organized as follows. Section \ref{sec:related} presents the related work on drone path finding and planning. 
Section \ref{sec:proposed} introduces the proposed Eco-flight path finding algorithm.
Section \ref{sec:results} compares EcoFlight with state-of-the-art path finding algorithms.
Section \ref{sec:conclusion} concludes the paper.

\section{Related Work}
\label{sec:related}

Although drone path planning has advanced significantly, several critical challenges remain. Many current methods rely on pre-planned routes and static environmental data, limiting their ability to adapt to dynamic changes such as moving obstacles, shifting wind conditions, or declining battery levels. Traditional path-finding methods like A* and Dijkstra’s algorithms perform well in stable environments but struggle in real-time scenarios \cite{wu2020adaptive, shivgan2020energy}. Some recent approaches adjust path weights dynamically \cite{bekmurzaeva2024application}, yet they still fall short when faced with unexpected situations.

Energy estimation presents another challenge. Most models assume that energy consumption depends solely on distance traveled \cite{shivgan2020energy}, overlooking factors such as acceleration, deceleration, and altitude changes. Some studies incorporate wind resistance and payload variable \cite{guganpath}, but these comprehensive models often require substantial computing power, limiting their practicality for real-time applications.

Artificial Intelligence (AI) is also being adopted to enhance drone navigation. Techniques such as reinforcement learning and deep learning allow drones to learn from experience and optimize flight paths over time \cite{li2018universal}. Although promising, these methods require extensive training and high computational resources. Hybrid approaches that combine traditional path planning with AI-based learning \cite{li2018universal} may balance efficiency and adaptability, but they still require further testing in diverse environments.

A major barrier to progress is the lack of a standardized framework for evaluating and comparing path-planning methods for moderately long flights between a single origin and destination, and particularly one with obstacles. Realistic flights may encounter obstacles that must be circumvented, yet research often relies on assuming flights on straight lines, with simple flight and energy models. However, efficient flights may be performed at low altitude. Therefore, the presence of obstacles on the flight path is a real possibility. Although existing work has proposed testing environments \cite{Ma2024}, a widely accepted standard has yet to emerge.

Addressing these challenges is essential for making drone path planning more realistic, reliable, and efficient. Other considerations are 
%Future research priorities include improving adaptability to real-time changes, enhancing multi-drone coordination, and 
incorporating more detailed energy models~\cite{hong2020least}. Establishing a standardized evaluation framework, especially for those with obstacles in point-to-point flights, would further increase accuracy and serve as a step towards flight automation.

\section{EcoFlight Algorithm}
\label{sec:proposed} 

The proposed EcoFlight algorithm incorporates an energy model in path-planning decisions in 3D space, and it is described next.
%This section describes EcoFlight, a {\it Rise and Traverse} algorithm used for comparison, and the methods of calculation the energy consumption used for the path energy estimation. 

\subsection{EcoFlight Algorithm}
%We use an approximation method to evaluate the difference between the predicted energy cost of a path and the remaining energy required. The A* algorithm then selects the most energy-efficient path by analyzing specific nodes along the path. A region for flying is represented by nodes separated by 1 m of distance and established as a three-dimensional topographic map indicating the height of each node. By incorporating an energy-based scaling factor, the algorithm prioritizes routes that minimize energy consumption. Vertical movement costs are calculated based on the drone's mass and the constant force of gravity, with higher energy penalties for climbing. In contrast, horizontal movement costs are influenced by resistance factors such as wind or drag. This method enables drones to identify paths that consume the least amount of energy, making it especially suitable for autonomous operation. 

Ecoflight uses an approximation method to evaluate the difference between the predicted energy cost of a path and the remaining energy required. It applies the A* algorithm to select the most energy-efficient route by analyzing the nodes on a 3D map. 
The flight region is represented as a 3D topographic map, with nodes spaced 1 meter apart and annotated with their elevation. An energy-based scaling factor prioritizes routes that minimize consumption: vertical movement costs are calculated from the drone’s mass and gravitational force, with higher penalties for climbing, while horizontal movement costs account for resistance factors such as wind or drag. This approach enables drones to select paths with minimal energy usage, making it well-suited for autonomous operation.

Algorithm~\ref{algo:Astar} presents the pseudocode for EcoFlight. Initially, EcoFlight creates two collections: OpenSet, containing nodes that have been discovered but have yet not fully evaluated, and ClosedSet, containing nodes already processed. The algorithm begins by setting $g$(start)=0 (the energy cost from the start to self) and computing $f$(start) = $g$(start) + $h$(start, goal), where $h$ estimates the remaining cost to the destination. During execution, the algorithm repeatedly selects the node in OpenSet with the lowest $f$-value, identifying it as the most promising candidate for an energy-efficient route. If the current node is the goal, the algorithm reconstructs the path following the parent pointers recorded during traversal and returns this path. Otherwise, the current node is removed from OpenSet, added to ClosedSet, and its neighbors are evaluated.

If a neighbor is already in ClosedSet, it is skipped. For each remaining neighbor, a tentative cost is calculated via the current node. If this cost is lower than any previously recorded cost, or if the neighbor has not yet been discovered, its cost values are updated:$g$(neighbor) is set to tentative cost, $f$(neighbor) is represented as $g$(neighbor)+$h$(neighbor, goal).
The parent pointer is set to the current node. If the neighbor is not in OpenSet, it is added for future evaluation. This process repeats until the goal is reached, in which case the path is reconstructed, or until OpenSet is empty, indicating that no valid path exists.

% \begin{algorithm}
% \caption{EcoFlight Algorithm}
% \label{algo:Astar}
% \KwIn{Start node $start$, End node $goal$}
% \KwOut{Optimal path from $start$ to $goal$}

% OpenSet $\gets \{start\}$\;
% ClosedSet $\gets \emptyset$\;
% $g(start) \gets 0$\;
% $f(start) \gets g(start) + h(start, goal)$\;

% \While{OpenSet is not empty}{
%     $current \gets$ node in OpenSet with lowest $f$ value\;
    
%     \If{$current = goal$}{
%         \Return reconstructed path from $current$\;
%     }
    
%     Remove $current$ from OpenSet\;
%     Add $current$ to ClosedSet\;
    
%     \ForEach{neighbor $n$ of $current$}{
%         \If{$n \in$ ClosedSet}{
%             \textbf{continue}\;
%         }
        
%         tentative\_g $\gets g(current) + cost(current, n)$\;
        
%         \If{$n \notin$ OpenSet \textbf{or} tentative\_g $<$ g(n)}{
%             $g(n) \gets$ tentative\_g\;
%             $f(n) \gets g(n) + h(n, goal)$\;
%             parent($n$) $\gets current$\;
            
%             \If{$n \notin$ OpenSet}{
%                 Add $n$ to OpenSet\;
%             }
%         }
%     }
% }
% \Return Failure (No Path Found)\;

% \end{algorithm}

\begin{algorithm}
\caption{EcoFlight Algorithm}
\label{algo:Astar}
\begin{algorithmic}[1]
\State \textbf{Input:} Start node $start$, Goal node $goal$
\State \textbf{Output:} Optimal path from $start$ to $goal$
\Statex
\State OpenSet $\gets \{start\}$
\State ClosedSet $\gets \emptyset$
\State $g(start) \gets 0$
\State $f(start) \gets g(start) + h(start, goal)$
\While{OpenSet is not empty}
    \State $current \gets$ node in OpenSet with lowest $f$ value
    \If{$current = goal$}
        \State \Return reconstructed path from $current$
    \EndIf
    \State Remove $current$ from OpenSet
    \State Add $current$ to ClosedSet
    \ForAll{neighbor $n$ of $current$}
        \If{$n \in$ ClosedSet}
            \State \textbf{continue}
        \EndIf
        \State tentative\_g $\gets g(current) + cost(current, n)$
        \If{$n \notin$ OpenSet \textbf{or} tentative\_g $<$ $g(n)$}
            \State $g(n) \gets$ tentative\_g
            \State $f(n) \gets g(n) + h(n, goal)$
            \State parent($n$) $\gets current$
            \If{$n \notin$ OpenSet}
                \State Add $n$ to OpenSet
            \EndIf
        \EndIf
    \EndFor
\EndWhile
\State \Return Failure (No Path Found)
\end{algorithmic}
\end{algorithm}

\subsection{Rise and Traverse Algorithm}
%The Rise and Traverse algorithm serves as a comparative baseline to evaluate the performance of EcoFlight. Algorithm~\ref{algo:RiseTraverse} presents the pseudocode for this obstacle avoidance strategy. Unlike EcoFlight, which dynamically searches for energy-efficient paths using heuristic cost functions, Rise and Traverse employs a segmented approach that divides navigation into two distinct phases. The first phase involves a vertical ascent to align the drone’s altitude with the goal’s final  $z$-coordinate. Once this ascent is complete, the drone proceeds with a horizontal traversal to reach the destination. If an obstacle is encountered, an A* search is triggered to find an alternate route, ensuring feasibility while maintaining a preference for upward movement. Although this method is computationally simpler, it lacks dynamic energy optimization and often produces suboptimal paths compared to EcoFlight. Specifically, the Rise and Traverse approach does not incorporate heuristic evaluations during flight and may result in additional energy consumption, especially noticeable in scenarios where diagonal movement would be more efficient. With this algorithm, we establish a performance benchmark that allows for a direct comparison. 
The Rise-and-Traverse algorithm serves as a baseline for comparing EcoFlight’s performance. %Algorithm~\ref{algo:RiseTraverse} presents the pseudocode for this obstacle avoidance strategy. 
Unlike EcoFlight, which dynamically searches for energy-efficient paths using heuristic cost functions, Rise and Traverse uses a segmented approach with two distinct phases. The first phase is a vertical ascent to match the drone’s altitude with the final destination’s $z$-coordinate. The second phase is a horizontal traversal to the destination. If an obstacle is encountered, an A* search is initiated to find an alternate route, ensuring feasibility while maintaining the preference for upward movement. While computationally simpler, this method lacks dynamic energy optimization and often produces suboptimal paths as compared to EcoFlight. Specifically, it does not incorporate heuristic evaluations for the flight, which can lead to increased energy consumption, particularly in scenarios where diagonal traveling would be more efficient.

\subsection{Energy Model}
%Accurate energy estimation is a critical component of autonomous path planning, particularly for missions requiring extended endurance or precise maneuvering. This section provides a detailed overview of the energy model used to evaluate drone energy consumption based on the path selected by each algorithm.
The energy model incorporates physical parameters such as drone mass, gravitational forces, and aerodynamic drag to calculate energy expenditure along each segment of a flight path using a force vector-based approach. %The primary objective is to validate the accuracy and effectiveness of energy-optimized A* path-finding algorithms.
%UNITS _____________________________________________ 
The energy consumption model in the simulator uses gravitational and kinematics equations that describe the energy consumption on drone flight dynamics.
%\subsubsection{Parameters}
Table~\ref{tab:parameters} shows the variables of the physical parameters used to model the energy expenditure estimation. %The constants were determined through experimental data obtained from a chassis test in our laboratory. 
\begin{table}[h]
\centering
\caption{Terms and Physical Parameters.}
\label{tab:parameters}
\renewcommand{\arraystretch}{1.2}  % Adjust row height for better readability
%\begin{tabular}{|>{\columncolor{gray!20}}l|c|c|}
\begin{tabular}{|l|c|c|}
\hline
\textbf{Parameter} & \textbf{Symbol} & \textbf{Unit} \\
\hline
Mass of drone & $m$ & \text{ $kg$} \\
\hline
Gravitational acceleration & $g$ & \text{m/s}$^2$ \\
\hline
Air density & $\rho$ & \text{ kg/m}$^3$ \\
\hline
Drag coefficient & $C_d$ & Dimensionless \\
\hline
Cross-sectional area & $A$ & \text{ m}$^2$ \\
\hline
Cruise speed & $v_c$ & \text{m/s} \\
\hline
Acceleration & $a$ & \text{m/s}$^2$ \\
\hline
\end{tabular}
\end{table}

\begin{figure}[htbp!]        % h = here, t = top, b = bottom, p = page
    \centering
    \includegraphics[width=0.42\textwidth]{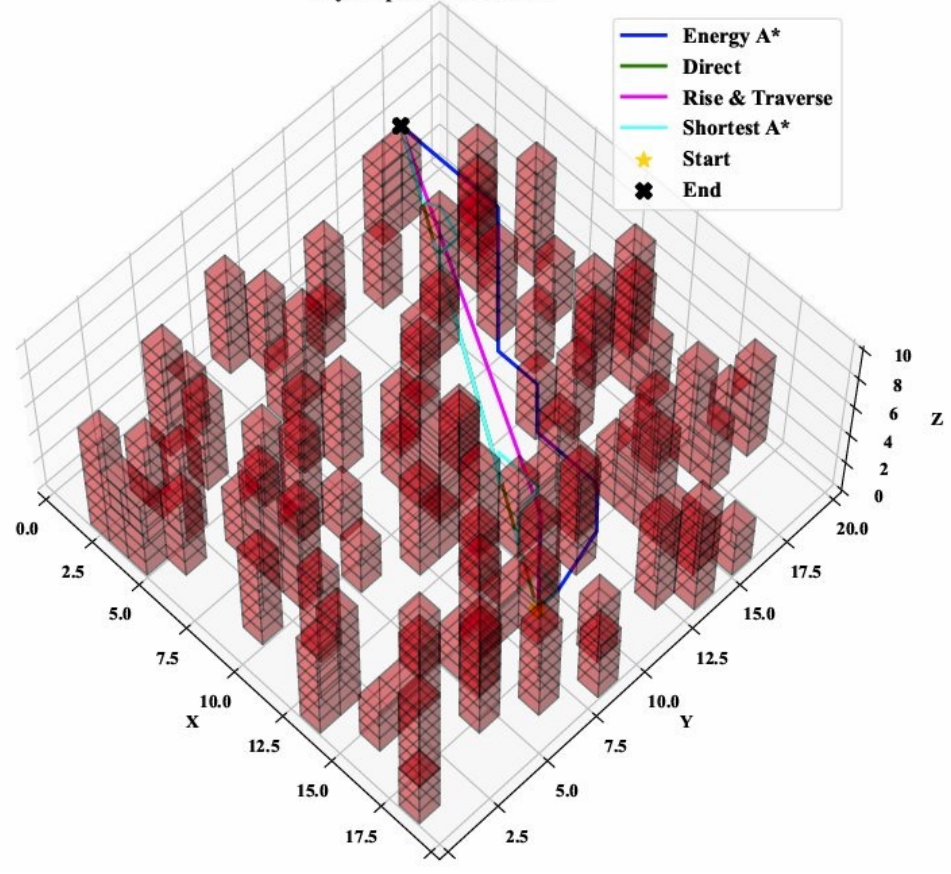}
    \caption{Sample of an area with obstacles of various heights and the paths followed by the compared path-finding algorithms.}
    \label{fig:obstacles}
\end{figure}

%FORCECALC _____________________________________________

%\subsubsection{Force Vector Calculations}

%The force vectors are defined as follows:

%\paragraph{
Gravity, thrust, and drag ~\cite{Anderson2010} forces are defined in (\ref{eq:gravity}), (\ref{eq:thrust}), and (\ref{eq:drag}), respectively.
\begin{equation}
\vec{F_g} = -mg
\label{eq:gravity}
\end{equation}

\begin{equation}
\vec{F_t} = m\vec{a}, \quad \vec{a} = \frac{\vec{v_{\text{new}}} - \vec{v_{\text{old}}}}{\Delta t}
\label{eq:thrust}
\end{equation}
\begin{equation}
F_d = \frac{1}{2} C_d A \rho v^2, \quad \vec{F_d} = -F_d \frac{\vec{v}}{|\vec{v}|}
\label{eq:drag}
\end{equation}

%\begin{center}
When hovering, \( \vec{a} \) is an upward acceleration equal to \( \vec{g} \).
%\end{center}

%\paragraph{Net Force} 
The {\it Net Force} is $\vec{F_{\text{net}}} = \vec{F_g} + \vec{F_d} + \vec{F_t}$.
% $\begin{equation}
% \vec{F_{\text{net}}} = \vec{F_g} + \vec{F_d} + \vec{F_t}
% \end{equation}

%ENERGYCALC _____________________________________________

\subsubsection{Energy Calculations}
The energy required is calculated as:
%\paragraph{
the work done by the net force, hovering energy~\cite{Halliday2013}, drag energy~\cite{Munson2013}, and acceleration energy~\cite{Tipler2007} are defined in (\ref{eq:work}), (\ref{eq:e_h}), (\ref{eq:e_d}), and (\ref{eq:e_a}), respectively.
\begin{equation}
W = \vec{F_{\text{net}}} \cdot \vec{d}, \quad \vec{d} = \vec{p_{\text{new}}} - \vec{p_{\text{old}}},
\label{eq:work}
\end{equation}

\begin{equation}
E_h = mgd_t,
\label{eq:e_h}
\end{equation}

\begin{equation}
E_d = F_d \cdot d,
\label{eq:e_d}
\end{equation}

\begin{equation}
E_a = \frac{1}{2} m (v_{\text{new}}^2 - v_{\text{old}}^2),
\label{eq:e_a}
\end{equation}
%\begin{center}
where \( d_t \) is the hovering time, as a function of distance and velocity. The energy consumed in each section of the path, or segment, is calculated:
%\end{center}
%\paragraph{Total Segment Energy}
\noindent Total energy for Segment $i$ is then:
\begin{equation}
E_{s_i} = E_h + E_d + E_a
\end{equation}

%\subsubsection{Total Path Energy}
The total energy for a path consisting of \( N \) segments \cite{Fossen2011} is:
\begin{equation}
E_{\text{total}} = \sum_{i=0}^{N-1} E_{s_i}
\end{equation}

%\subsection{Path Energy Application}
The path energy calculations presented here form the basis for directly comparing different flight paths in terms of energy efficiency. By quantifying the energy expenditure for each path, this approach enables a direct evaluation of EcoFlight’s efficiency and effectiveness relative to alternative algorithms. The scheme ensures that path selection prioritizes minimal energy consumption while preserving optimal traversal performance.

\begin{figure*}[htb!]
    \centering
    % First row (2 columns)
    \subcaptionbox{30\% obstacle density at 0.3 m/s.
    \label{figure:EcoFlight1}}
    {%
     \includegraphics[width=0.45\textwidth]{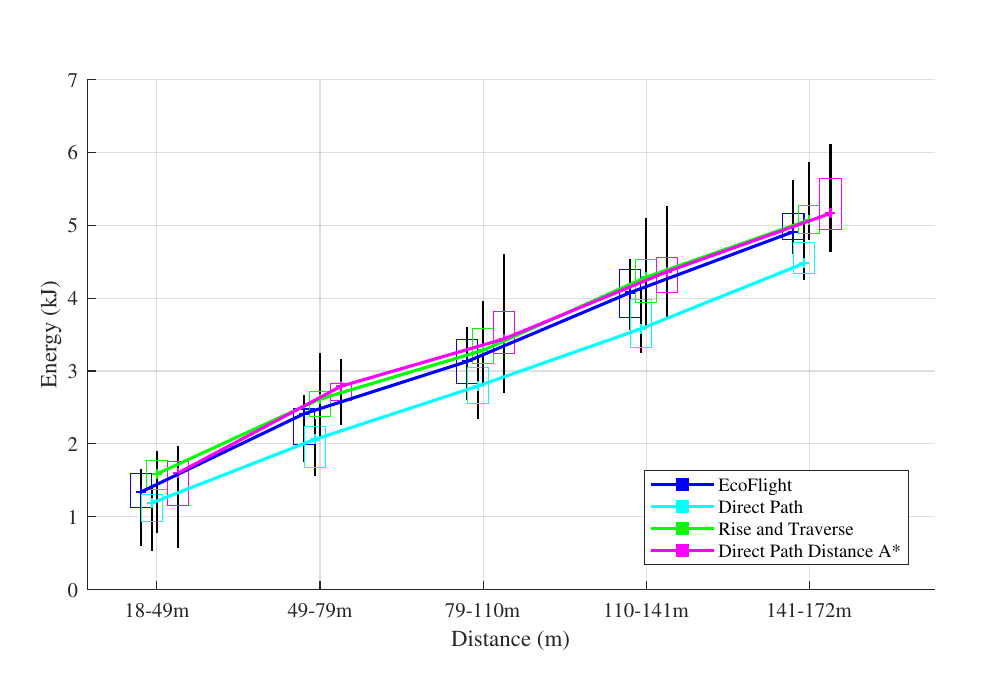}
    }
    \subcaptionbox{30\% obstacle density at 3 m/s.
    \label{figure:EcoFlight2}}
    {
     \includegraphics[width=0.45\textwidth]{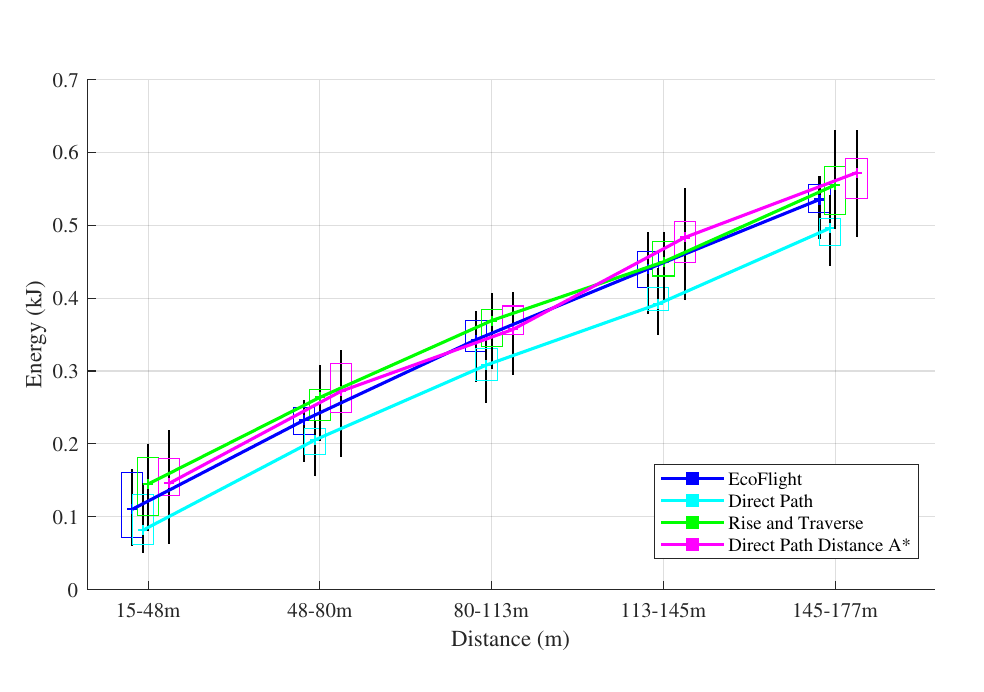}
    }
 %end 250502
    
    % \begin{subfigure}[b]{0.45\textwidth}
    %     \centering
    %     \includegraphics[width=\textwidth]{figures/30DS.pdf}
    %     \caption{30\% Building Density at .3 m/s}
    %     \label{figure:EcoFlight1}
    % \end{subfigure}
    % \hfill
    % \begin{subfigure}[b]{0.45\textwidth}
    %     \centering
    %     \includegraphics[width=\textwidth]{figures/30DF.pdf}
    %     \caption{30\% Building Density at 3 m/s}
    %     \label{figure:EcoFlight2}
    % \end{subfigure}

 \vspace{1em} % Vertical space between rows

\subcaptionbox{50\% obstacle density at 0.3 m/s.
    \label{figure:EcoFlight3}}
    {%
    % What you would typically have here is something like:
     \includegraphics[width=0.45\textwidth]{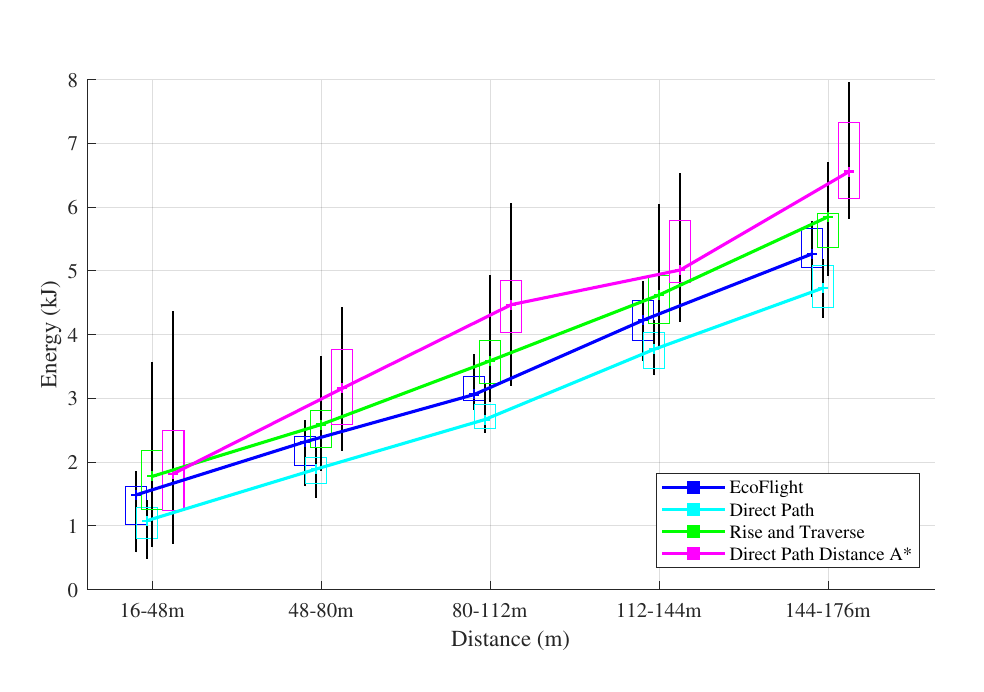}
    }
    \subcaptionbox{50\% obstacle density at 3 m/s.
    \label{figure:EcoFlight4}}
    {%
    % What you would typically have here is something like:
     \includegraphics[width=0.45\textwidth]{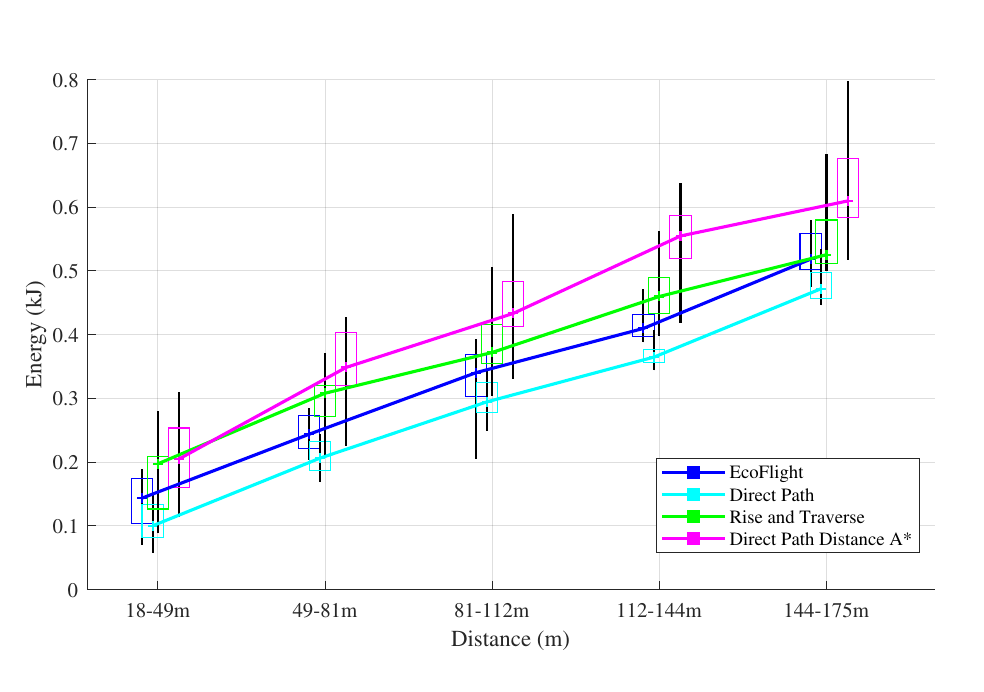}
    }
    
    % Second row (2 columns)
    % \begin{subfigure}[b]{0.45\textwidth}
    %     \centering
    %     \includegraphics[width=\textwidth]{figures/50DS.pdf}
    %     \caption{50\% Building Density at .3 m/s}
    %     \label{figure:EcoFlight3}
    % \end{subfigure}
    % \hfill
    % \begin{subfigure}[b]{0.45\textwidth}
    %     \centering
    %     \includegraphics[width=\textwidth]{figures/50DF.pdf}
    %     \caption{50\% Building Density at 3 m/s}
    %     \label{fig:EcoFlight4}
    % \end{subfigure}

    \vspace{1em} % Vertical space between rows
    
\subcaptionbox{75\% obstacle density at 0.3 m/s.
    \label{fig:EcoFlight5}}
    {%
    % What you would typically have here is something like:
     \includegraphics[width=0.45\textwidth]{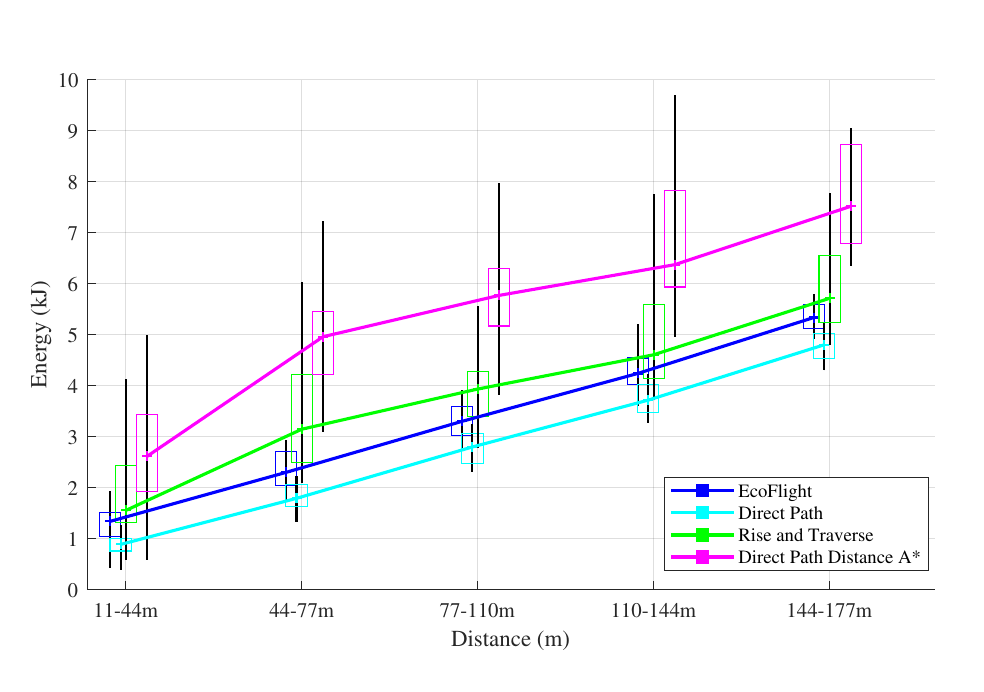}
    }
    \subcaptionbox{75\% obstacle density at 3 m/s.
    \label{fig:EcoFlight6}}
    {%
    % What you would typically have here is something like:
     \includegraphics[width=0.45\textwidth]{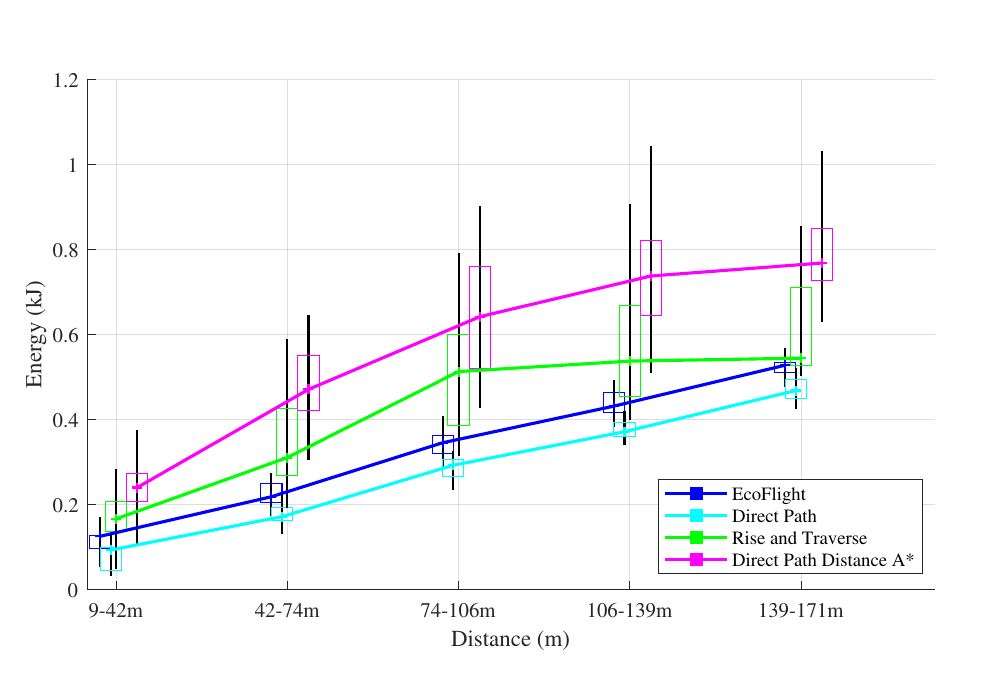}
    }
    % Third row (2 columns)
    % \begin{subfigure}[b]{0.45\textwidth}
    %     \centering
    %     \includegraphics[width=\textwidth]{figures/75DS.pdf}
    %     \caption{75\% Building Density at .3 m/s}
    %     \label{fig:EcoFlight5}
    % \end{subfigure}
    % \hfill
    % \begin{subfigure}[b]{0.45\textwidth}
    %     \centering
    %     \includegraphics[width=\textwidth]{figures/75DF.pdf}
    %     \caption{75\% Building Density at 3 m/s}
    %     \label{fig:EcoFlight6}
    % \end{subfigure}

    \caption{Comparison of energy consumption of EcoFlight to other path-planning algorithms.}
    \label{figure:EcoFlightGrid2x3}
\end{figure*}

\section{Evaluation}
\label{sec:results}

%\subsection{Flight path and energy consumption}
%\begin{figure}[h!]        % h = here, t = top, b = %bottom, p = page
%    \centering
%    \includegraphics[width=0.5\textwidth]{figures/pointmassGraphs-cropped-1-1.pdf}
%    \caption{Flight dynamics on the calibration path(s).}%Figure ~\ref{MLgraphic}.}
%    \label{MLgraphs}
%\end{figure}

%This section demonstrates EcoFlight's performance in navigating simulated test environments. 
%By generating a variety of scenarios, 
We evaluated EcoFlight's ability to identify energy-efficient paths and compare its energy consumption to that of other path planning methods under various obstacle density environments for calibration and testing.% using Matlab. %The primary objective is to analyze energy usage along the computed paths, thereby validating the algorithm’s effectiveness for real-world applications, particularly in autonomous drone operations where energy efficiency is critical.
EcoFlight selects an energy-efficient free-obstacle path. Direct Path ignores obstacles entirely, moving in a straight line from start to goal. Direct Path Distance A* follows the shortest path until it encounters an obstacle, at which point it runs a localized and unadjusted A* search to find the shortest detour. Rise and Traverse first moves along the $z$-axis to match the destination altitude, then proceeds linearly to the goal while avoiding obstacles, similar to Direct Path Distance.

We tested the algorithms under three randomly generated obstacles, with average densities of 30\%, 50\%, and 75\%, and with flights at two different speeds: 0.3 m/s and 3 m/s. Due to the high computational cost, each configuration was limited to 100 trials. In each trial, the origin and destination were randomly selected within the test area, ensuring the drone maintained a flight altitude at least one unit above any obstacle. Figure \ref{fig:obstacles} shows an example of an area with 30\% of the volume with obstacles. The location of the obstacles and their height are randomly selected according to the average density. The drone may fly around or above obstacles through the most energy-efficient path. This figure also shows a glimpse of the possible paths selected by the algorithms used for comparison.

EcoFlight demonstrated remarkable improvements as obstacle density increased. High-density environments prompted the algorithm to generate smoother and more energy-efficient paths, though at the cost of increased computation time.

As a benchmark, the Direct Path method was the most energy-efficient because this approach ignores the obstacles entirely, cutting straight through them along the shortest possible path. Among the realistic obstacle-aware methods, EcoFlight consistently produced the lowest-energy routes on average. Figures~\ref{figure:EcoFlight1} and~\ref{figure:EcoFlight2} illustrate the average energy consumption in a simulated environment with 30\% obstacle density. With relatively few obstacles, most algorithms produce similar routes and thus exhibit comparable energy usage, forming a tight cluster.
However, as obstacle density increases, as seen in Figures~\ref{figure:EcoFlight3} and~\ref{figure:EcoFlight4}, the need to navigate around obstacles becomes more pronounced, and EcoFlight’s advantage becomes clearer. These results also highlight the consistently higher energy cost incurred by Direct Path Distance A* due to its frequent, reactive detours. This trend is further accentuated in Figures~\ref{fig:EcoFlight5} and~\ref{fig:EcoFlight6}, which correspond to the highest obstacle density tested.

\section{Conclusions}
\label{sec:conclusion}
In this paper, we propose EcoFlight, an energy-efficient path planning algorithm that selects energy-efficient flight paths while avoiding obstacles in a 3D environment. We evaluated EcoFlight in environments with varying obstacle densities and compared it with Direct Path, Rise and Traverse, and Direct Path Distance A* algorithms. 
%distance-based A*, Rise-and-Traverse obstacle avoidance, and obstacle-agnostic direct pathfinding algorithms. 
The simulation results show that EcoFlight outperforms the other obstacle-aware algorithms in terms of energy consumption. %, especially at high density obstacle environment. 
%Such obstacles represent actual scenarios, such as city buildings or tree canopies.
%High density obstacles environment provides realistic scenarios in urban settings and forests with various-height tree canopies. 
This advantage becomes more significant as obstacle density in the fly zone increases, indicating that some fly maneuvers may be more costly than others and that distance itself may not be proportional to the energy expenditure in the general case. These obstacle scenarios represent spaces of densely populated urban areas with tall buildings. However, the modeling of tree canopies may require a more complex obstacle model to make it realistic.

%--------------------------
%--------------------------
%--------------------------

%\section*{Acknowledgment}
%\ifCLASSOPTIONcompsoc
 % \section*{Acknowledgments}
%\else
\section*{Acknowledgment}
%\fi
This material is based upon work partially supported by the National Science Foundation under Grant Awards 1856032 and 2402240 and NJIT Faculty Seed Grant. 

\bibliographystyle{IEEEtran}
%\bibliography{references}

%\section*{References}

% Please number citations consecutively within brackets \cite{b1}. The 
% sentence punctuation follows the bracket \cite{b2}. Refer simply to the reference 
% number, as in \cite{b3}---do not use ``Ref. \cite{b3}'' or ``reference \cite{b3}'' except at 
% the beginning of a sentence: ``Reference \cite{b3} was the first $\ldots$''

% Number footnotes separately in superscripts. Place the actual footnote at 
% the bottom of the column in which it was cited. Do not put footnotes in the 
% abstract or reference list. Use letters for table footnotes.

% Unless there are six authors or more give all authors' names; do not use 
% ``et al.''. Papers that have not been published, even if they have been 
% submitted for publication, should be cited as ``unpublished'' \cite{b4}. Papers 
% that have been accepted for publication should be cited as ``in press'' \cite{b5}. 
% Capitalize only the first word in a paper title, except for proper nouns and 
% element symbols.

% For papers published in translation journals, please give the English 
% citation first, followed by the original foreign-language citation \cite{b6}.

\end{document}